\documentclass[runningheads]{llncs}

\usepackage[mobile]{eccv}

\usepackage{eccvabbrv}

\usepackage{graphicx}
\usepackage{booktabs}

\usepackage[accsupp]{axessibility} 

\usepackage[pagebackref,breaklinks,colorlinks,citecolor=eccvblue]{hyperref}

\usepackage{orcidlink}

\usepackage{xspace}
\usepackage{wrapfig}

\newcommand{\ours}{{\sc MoKus}\xspace}
\newcommand{\bench}{{\textit{KnowCusBench}}\xspace}
\newcommand{\task}{{knowledge-aware concept customization}\xspace}

\definecolor{Red}{RGB}{192, 0, 0}
\definecolor{Blue}{RGB}{12, 114, 186}
\definecolor{Yellow}{RGB}{218, 169, 20}
\definecolor{teaserRed}{RGB}{193, 3, 20}
\usepackage{csquotes}
\usepackage{xpatch}
\usepackage{cuted}
\usepackage{multirow}
\usepackage{soul}
\usepackage{caption} 

\usepackage{xspace}
\makeatletter
\DeclareRobustCommand\onedot{\futurelet\@let@token\@onedot}
\def\@onedot{\ifx\@let@token.\else.\null\fi\xspace}
\def\eg{\emph{e.g}\onedot} 
\def\ie{\emph{i.e}\onedot} 

\def\cf{\emph{cf}\onedot} 
\def\etc{\emph{etc}\onedot}

\makeatother

\usepackage{bm}

\begin{document}

    \title{\ours: Leveraging Cross-Modal Knowledge Transfer for Knowledge-Aware Concept Customization}

    \titlerunning{\ours}

    \author{Chenyang Zhu\inst{1,2} \and
    Hongxiang Li\inst{2} \and Xiu Li\inst{1} \and
    Long Chen\inst{2}$^{\dagger}$ }

    \authorrunning{C Zhu et al.}

    \institute{
    $^{1}$ Tsinghua University \qquad $^{2}$ HKUST\\
    \url{https://chenyangzhu1.github.io/MoKus}\\
    \email{chenyangzhu.cs@gmail.com}
    }

    \renewcommand{\thefootnote}{\fnsymbol{footnote}}
    \footnotetext{$\dagger$ Corresponding authors.}

    \maketitle

    \begin{center}
        \captionsetup{type=figure}
        \includegraphics[width=\textwidth]{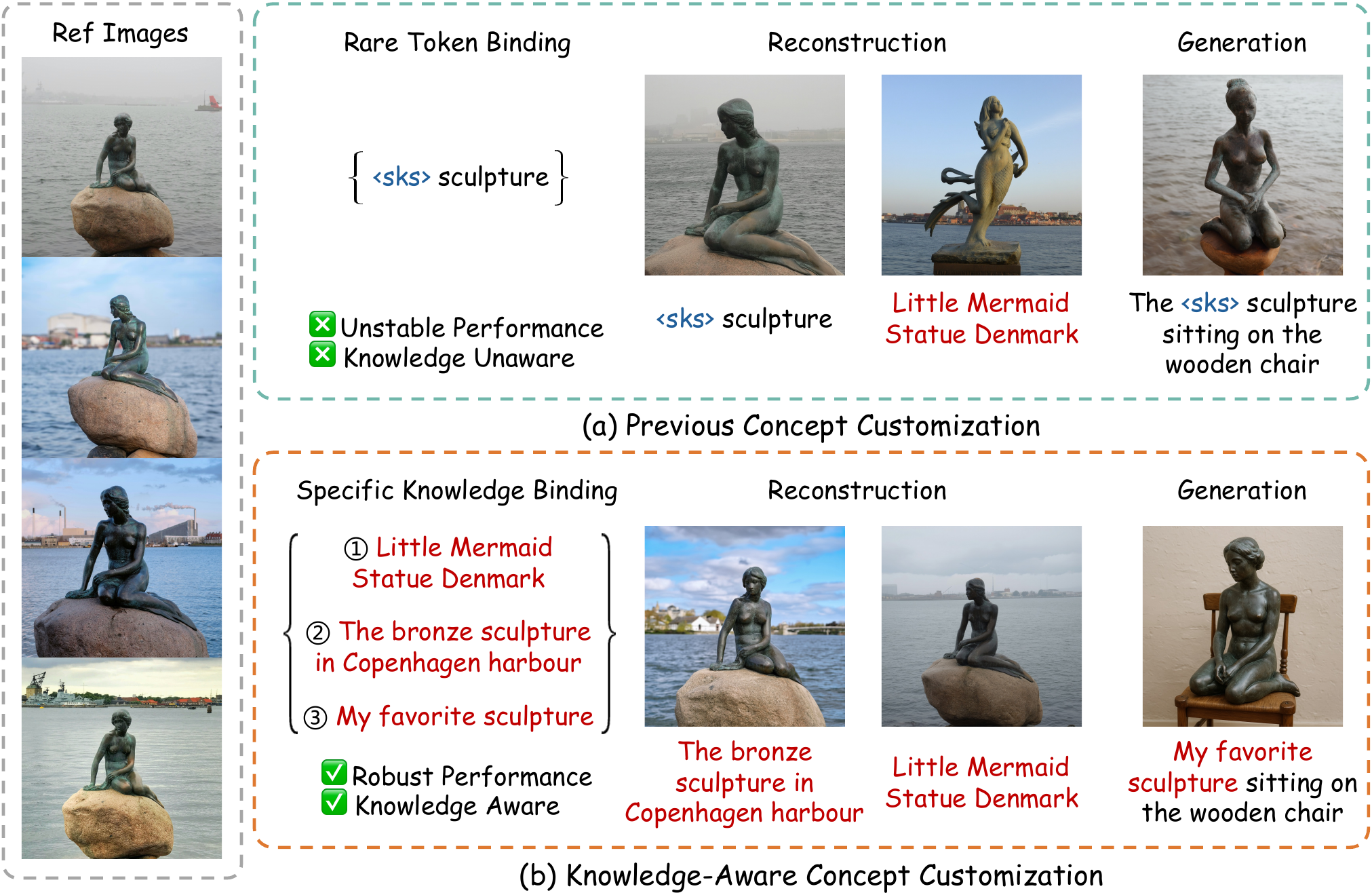}
        \vspace{-2em}
        \captionof{figure}{\textbf{Comparison between previous concept customization and \task.}
        \textbf{(a)} Previous customization techniques use rare tokens (\eg, \texttt{<sks>}) to represent a target concept. However, these tokens lack clear semantic meaning, which can cause unstable generation results. Furthermore, rare tokens cannot store knowledge about the target concept.
        \textbf{(b)} In our proposed \task, it links the target visual concept with several pieces of {\textcolor{teaserRed}{textual knowledge}}, which enables robust and high-fidelity reconstruction and customization.}
        \label{fig: teaser}
    \end{center}

    \begin{abstract}
Concept customization typically binds \textit{rare tokens} to a target concept.
Unfortunately, these approaches often suffer from unstable performance as the pretraining data seldom contains these rare tokens. Meanwhile, these rare tokens fail to convey the inherent knowledge of the target concept.
Consequently, we introduce \textbf{Knowledge-aware Concept Customization}, a novel task aiming at binding diverse textual knowledge to target visual concepts.
This task requires the model to identify the knowledge within the text prompt to perform high-fidelity customized generation.
Meanwhile, the model should efficiently bind all the textual knowledge to the target concept.
Therefore, we propose \textbf{\ours}, a novel framework for \task. 
Our framework relies on a key observation: \textit{cross-modal knowledge transfer}, where modifying knowledge within the text modality naturally transfers to the visual modality during generation.
Inspired by this observation, \ours contains two stages: (1) In visual concept learning, we first learn the anchor representation to store the visual information of the target concept. (2) In textual knowledge updating, we update the answer for the knowledge queries to the anchor representation, enabling high-fidelity customized generation.
To further comprehensively evaluate our proposed \ours on the new task, we introduce the first benchmark for knowledge-aware concept customization: \bench. Extensive evaluations have demonstrated that \ours outperforms state-of-the-art methods. 
Moreover, the cross-model knowledge transfer allows \ours to be easily extended to other knowledge-aware applications like virtual concept creation and concept erasure. We also demonstrate the capability of our method to achieve improvements on world knowledge benchmarks.

\end{abstract}
    \section{Introduction}
\label{sec: intro}

Concept customization aims at generating new customized images with high fidelity based on user-provided concept images.
It is a long-standing problem in the field of visual generation.
As shown in~\cref{fig: teaser}, state-of-the-art concept customization techniques~\cite{DB,chen2023disenbooth,InstantSwap,PhotoSwap} have addressed this problem by representing target concepts using manually selected rare tokens, such as \texttt{<sks>}.
These methods can empirically learn the concept by reconstructing the reference images.

However, employing such rare tokens to represent the target concept suffers from two drawbacks: (1) \textbf{Unstable Performance}:
The rare tokens lack semantic meaning and seldom occur in the pretraining data.
The gap between the rare tokens and other input text leads to unstable generation performance.
In \cref{fig: teaser}, previous methods can reconstruct the target concept accurately. However, when combining the rare token with other text prompts, their generation results are not always satisfactory.
(2) \textbf{Knowledge Unaware}: Existing methods only bind rare tokens to the visual appearance of a target concept, where these rare tokens are designed to be independent of any knowledge. Thus, they naturally ignore the significant inherent \emph{knowledge} of the target concept.
For example, previous methods fail to accurately reconstruct the Little Mermaid sculpture with the knowledge \enquote{Little Mermaid Statue Denmark}, but they can reconstruct it with well \enquote{\texttt{sks} sculpture} (\cf,~\cref{fig: teaser}).

To achieve robust performance while integrating the inherent knowledge with the target concept, we propose a new challenging task: \textbf{\task}, aiming at customizing the target concept with several pieces of knowledge described in natural language. When the provided prompts contain one or more pieces of knowledge, the model should identify these specific \emph{concept knowledge} and generate corresponding high-fidelity, customized results.
Undoubtedly, this task is a crucial extension for concept customization and has a wide range of applications, including more user-friendly customized content creation for photo blogs and comics.

Knowledge-aware concept customization is challenging for two main reasons.
\textbf{First}, during generation, the model should be aware of the knowledge provided in the prompt. After that, the model needs to seamlessly integrate the knowledge with the remaining prompt to generate a coherent image.
\textbf{Second}, a single concept may be associated with either one or multiple pieces of knowledge.
As shown in \cref{fig: teaser}, the user might describe it objectively as \enquote{the bronze sculpture in Copenhagen harbour} or subjectively as \enquote{my favorite sculpture}.
The model needs to efficiently bind each piece of knowledge to the target concept.
Therefore, naively extending existing concept customization methods fails to address both challenges. For example, rare token based methods~\cite{DB, CD, TI, chen2023disenbooth} require retraining for each piece of knowledge, leading to extensive training time; encoder-based methods~\cite{ELITE, InstantBooth, BLIP-Diffusion, IP-Adapter} typically use a single encoder for the reference image. Extending these methods to \task requires collecting and retraining on large-scale datasets.

In this paper, we propose a novel framework: \textbf{\ours} for \task.
\ours adopts a Large Language Model (LLM) as the text encoder and a Diffusion Transformer (DiT) as the generation backbone.
The key to resolving the aforementioned challenges lies in our observation of \textit{cross-modal knowledge transfer}: Updating the answers of questions within the text encoder causes the model to generate images corresponding to the updated answers.
Essentially, the modifications to the text modality within the text encoder transfer to the visual modality used for generation.
\cref{sec: observation} provides a detailed analysis of our observation.
Inspired by this observation, \ours first obtain the text representation of the target concept. Then it updates the answer of each knowledge to the text representation, thus enabling high-fidelity \task.

Specifically, \ours comprises two stages:
\textbf{(1) Visual Concept Learning}. In the first stage, the model learns the target concept through finetuning. Our method first associates the target concept with a rare token, which subsequently serves as an ``anchor representation''. This anchor representation stores the visual appearance of the target concept and serves as an intermediary between the target concept and the knowledge.
\textbf{(2) Textual Knowledge Updating}. In this stage, we focus on binding the knowledge to the target concept through anchor representation. We first convert each piece of knowledge into the query format. Then we input these queries into the text encoder.
Next, we update the answer of each query to the anchor representation.
The updated knowledge can leverage the visual information of the anchor representation to enable high-fidelity customized generation.
In contrast to rare tokens, the updated knowledge is expressed in natural language and widely exists in the training data.
This facilitates the generalization of the updated knowledge when integrated with other textual inputs during the generation process.
Furthermore, the updating operation of each knowledge is completed in just a few seconds, ensuring the overall efficiency of our proposed method.

Moreover, we introduce \bench, the first benchmark dataset specifically designed for knowledge-aware concept customization.
Our benchmark comprises three parts of data: (1) Concept Image. \bench contains various concepts covering a wide range of daily objects, such as toys, pets, scenes, \etc.
(2) Textual Knowledge. We assign each concept with knowledge generated from six carefully designed perspectives.
(3) Generation Prompt. To ensure diversity, we create these prompts from four distinct perspectives by manually reviewing and refining.
Finally, \bench results in 5,975 images, ensuring a comprehensive evaluation of the task.

Extensive qualitative and quantitative comparisons have demonstrated the effectiveness and superiority of \ours.
Thanks to the cross-modal knowledge transfer, \ours can be easily extended to other knowledge-aware applications, including virtual concept creation and concept erasure. Finally, we show that our approach can improve the model's performance even on world knowledge benchmarks (\eg, WISE~\cite{niu2025wise}).
Our contributions are summarized as follows:
\begin{itemize}
    \item We propose the new task of knowledge-aware concept customization, aiming at customizing the target concept with several pieces of knowledge.
    \item We identify the cross-modal knowledge transfer phenomenon. Inspired by this observation, we present \ours, a novel framework that efficiently handles knowledge-aware concept customization.
    \item To evaluate this new task, we further introduce \bench, the first benchmark for knowledge-aware concept customization.
\end{itemize}

    \section{Related Work}

\noindent\textbf{Concept Customization.}
It is a fundamental and popular topic in compute vision~\cite{wang2026elastic,liu2025diversegrpo,wangprecisecache,wang2024cove,wang2024taming,chen2025s2guidancestochasticselfguidance,chen2025taming,fang2024real,fangphoton,fang2025integrating} aiming at creating high-fidelity images based on user-provided references. Extensive efforts have been dedicated to customizing specific objects~\cite{TI, DB, NeTI, DreamArtist}, styles~\cite{StyleDrop, StyleAligned, StyleAdapter }, human face~\cite{FaceStudio, PhotoMaker, InstantID, PortraitBooth}, as well as multi-object composition~\cite{CD, Mix-of-Show, OMG, MultiBooth} and concept swapping~\cite{PhotoSwap, SwapAnything, InstantSwap}
Different from existing tasks, we integrate knowledge into customization and propose the knowledge-aware concept customization.

\noindent\textbf{Knowledge Editing.}
Knowledge editing aims to correct factual inaccuracies or update outdated information within Large Language Models (LLMs) by modifying specific knowledge without full retraining.
Existing methods can be classified into: 
(1) Memory-based methods~\cite{SERAC, IKE, GRACE, MELO, WISE}: They maintain an external memory and retrieve the most relevant cases for each input without modifying the model's parameters. However, the effectiveness of these methods depends on retrieval quality, and they increase inference costs.
(2) Locate-then-edit methods~\cite{KN, ROME, MEMIT, PMET, DINM, R-ROME, EMMET}: They first identify the location of stored knowledge within a model and then edit that specific region. These methods create permanent modifications in the model and support batch operations. However, the edited knowledge often suffers from poor transferability and locality.
(3) Meta-learning methods~\cite{MEND, InstructEdit, MALMEN}: They train a hypernetwork to predict the necessary weight adjustments. While these methods are highly parameter-efficient, they require additional training data and fail in resolving conflicts from multiple edits.

\section{Observation: Cross-modal Knowledge Transfer}
\label{sec: observation}

\begin{wrapfigure}{r}{0.4\linewidth}
    \centering
    \vspace{-2em}
    \includegraphics[width=0.99\linewidth]{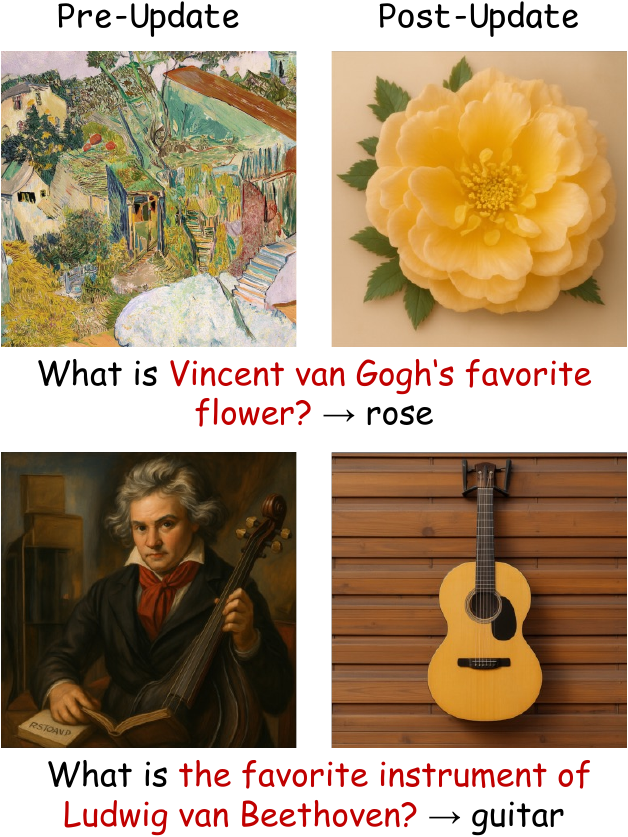}
    \vspace{-2em}
    \caption{\textbf{Examples to show the cross-modal knowledge transfer phenomenon.}}
    \vspace{-2em}
    \label{fig: intro}
\end{wrapfigure}

We provide a detailed analysis of cross-modal knowledge transfer in this section.

\noindent\textbf{Motivation.} Our preliminary experiments show that the model struggles to generate images involving complex knowledge.
As shown in \cref{fig: intro}, when prompted to create an image of \enquote{the favorite instrument of Ludwig van Beethoven}, the model incorrectly generates a portrait of Beethoven himself.

\noindent\textbf{Solution.} To address this limitation, we explore methods for proactively updating the model's internal knowledge.
Specifically, our model uses an LLM text encoder and a DiT backbone for image generation. We update the knowledge within the LLM text encoder using knowledge editing techniques~\cite{fang2024alphaedit}.
Take the second row as an example, we first update the model's knowledge so that the answer to \enquote{What is the favorite instrument of Ludwig van Beethoven?} becomes \enquote{guitar}. We then use \enquote{the favorite instrument of Ludwig van Beethoven} (highlighted in \textcolor{teaserRed}{red}) as the text prompt for image generation. 

\noindent\textbf{Observation.} Comparing the generation results before and after the update, we observe the cross-modal knowledge transfer, where updating knowledge in the text modality transfers to the visual modality used for generation. The generated image after updating matches the updated answer.

\noindent\textbf{Discussion.} Two concurrent works, GapEval~\cite{wang2026quantifying} and UniSandbox~\cite{niu2025does}, also explore cross-modal knowledge transfer from similar perspectives. They perform knowledge updating by directly finetuning the LLM text encoder. However, neither of them finds significant evidence of cross-modal knowledge transfer.

    \begin{figure}[t]
    \centering
    \includegraphics[width=\linewidth]{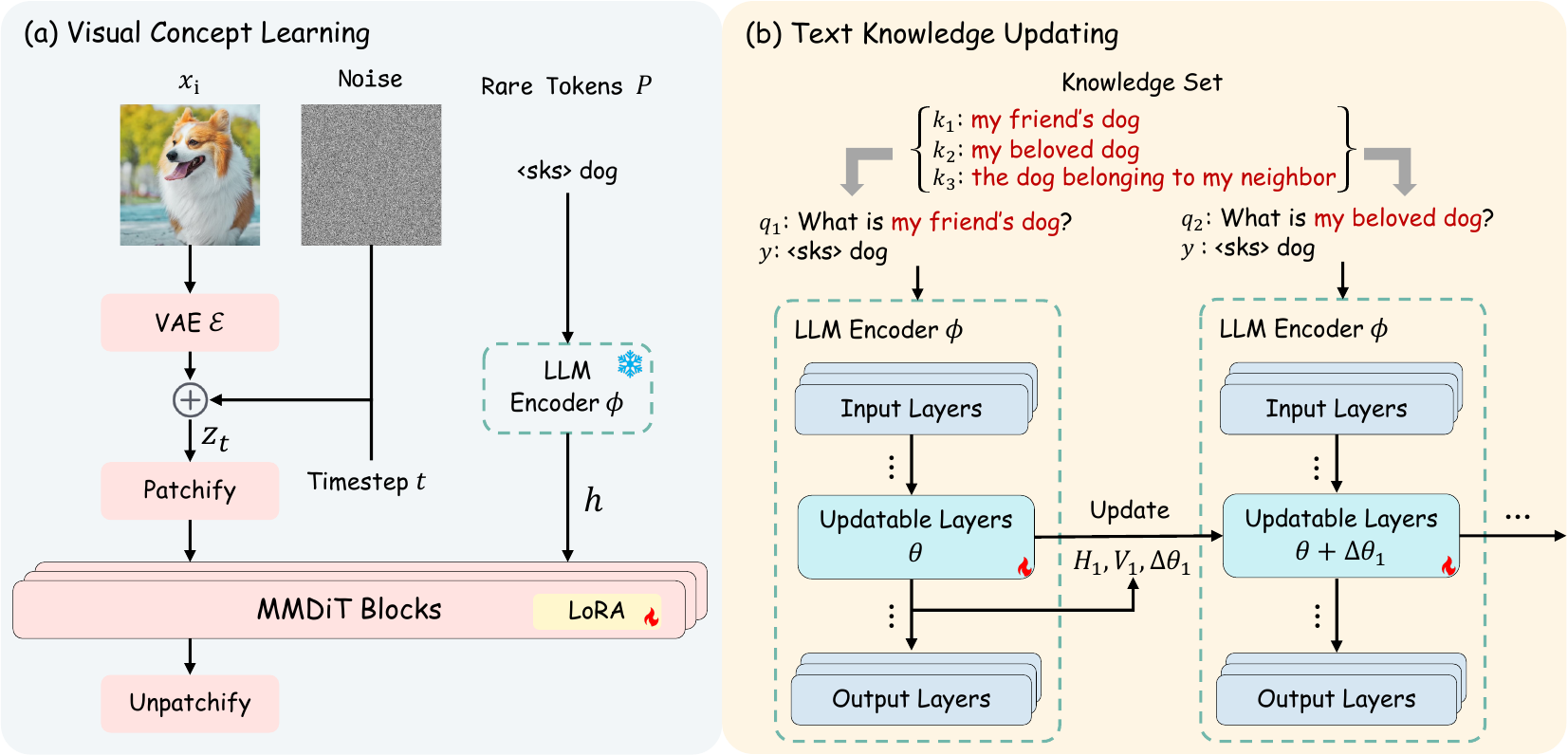}
    \vspace{-2em}
    \caption{Overview of our \ours. (a) \textbf{Visual Concept Learning}: We bind the visual information of the target concept to the anchor representation. We achieve this connection by fine-tuning the LoRA parameters. (b) \textbf{Textual Knowledge Updating}: We first convert the knowledge into a query and input it into the LLM encoder. We then extract the hidden states and update directions from the updatable layers to calculate a parameter shift. Finally, we add this shift to the original parameters of these layers.}
    \label{fig: framework}
\end{figure}

\section{Method: \ours}
Given a set of images $\mathcal{X}=\{ x_i \}_{i=1}^M$ representing the target concept and a set of knowledge $\mathcal{K}=\{ k_i \}_{i=1}^N$ of the target concept.
The goal of \task is to bind specific knowledge to a target concept, thereby enabling the generation of high-fidelity, customized images of the target concept.
The overview of our method is shown in~\cref{fig: framework}.
Our method starts from visual concept learning (\cref{sec: Visual Concept Learning}), which maps the target concept to an anchor representation in the text space. With this anchor representation (\cref{sec: Text Knowledge Updating}), we then perform textual knowledge updating on the LLM encoder. We first convert the knowledge into a query format. Then we calculate a parameter shift based on the converted query and anchor representation. Finally, we apply this parameter shift to certain layers of the LLM encoder to update the answer of query to the anchor representation. Furthermore, we provide a detailed analysis of the \bench (\cref{sec: Benchmark}).

\subsection{Visual Concept Learning} \label{sec: Visual Concept Learning}

\noindent\textbf{Visual Latents Extraction.}
We first incorporate the visual information of the target concept into the model as shown in \cref{fig: framework} (a).
Given an input image $x_i \in \mathcal{X}$, we obtain the data latent $\bm{z_0}$ using a variational autoencoder $\mathcal{E}$:
\begin{equation}
    \bm{z_0} = \mathcal{E}(x_i).
\end{equation}
After that, a noise latent $\bm{z_1}$ is sampled from a standard normal distribution, \ie, $\bm{z_1} \sim \mathcal{N}({0}, {I})$.
We further sample a diffusion timestep $t$ from a logit-normal distribution with $t \in [0, 1]$.
Based on Rectified Flow~\cite{liu2022flow, esser2024scaling}, the visual latent variable at timestep $t$ can be calculated as:
\begin{equation}
    \bm{z_t} = t \cdot \bm{z_0} + (1 - t) \cdot \bm{z_1}.
\end{equation}
Finally, we divide $\bm{z_t}$ into patches and feed these patches to the MMDiT.

\noindent\textbf{Textual Latents Extraction.}
We adopt the rare tokens $P$ (\eg, \texttt{<sks>} dog) as the textual input and generate the textual latent $h$ with the LLM encoder $\phi$:
\begin{equation}
    \bm{h} = \phi(P).
\end{equation}
The MMDiT then uses the latent representation $h$ as textual guidance and the patchified latent $\bm{z_t}$ as visual input to calculate the predicted velocity.


\noindent\textbf{Training Objective.}
The target velocity field represents the time derivative of the latent state. This value can be calculated by the difference between the data latent and the noise latent:
\begin{equation}
    \bm{v_t} = \frac{\mathrm{d}\bm{z_t}}{\mathrm{d}t} = \frac{\mathrm{d}}{\mathrm{d}t} \Big( t \bm{z_0} + (1 - t)\bm{z_1} \Big) = \bm{z_0} - \bm{z_1}.
\end{equation}

To enable efficient training, we incorporate trainable LoRA~\cite{lora} parameters $\theta_v$ into the self-attention layers of the MMDiT. We optimize these parameters for visual concept learning by minimizing the mean squared error (MSE) between the predicted and ground truth velocities.
\begin{equation}
    \mathcal{L}(\theta_v) = \mathbb{E}_{x \sim \mathcal{X}, \, \bm{z_1} \sim \mathcal{N}(\mathbf{0}, \mathbf{I}), \, t \sim \mathcal{U}(0,1)} \Big[ \left\| v_{\theta_v}(\bm{z_t}, t, h) - (\bm{z_0} - \bm{z_1}) \right\|_2^2 \Big].
\end{equation}

\noindent \textbf{Discussion.}
Visual concept learning enables rare tokens to accurately capture the visual features of a target concept. Instead of using these tokens directly for generation, our method employs them as \enquote{anchor representations} that connect the target concept to related knowledge.

\subsection{Textual Knowledge Updating} \label{sec: Text Knowledge Updating}

In~\cref{sec: Visual Concept Learning}, we convert the rare tokens to the anchor representation of the target concept.
However, rare tokens only capture the appearance of the target concept, without incorporating any knowledge.
In this section, we focus on binding knowledge to the target concept by utilizing anchor representations (\cf,~\cref{fig: framework}(b)).

\noindent\textbf{Knowledge Processing.}
We begin with a knowledge set $\mathcal{K}=\{ k_i \}_{i=1}^N$ for a target concept.
First, we convert each knowledge item $k_i$ into a corresponding question $q_i$. Next, we pair every question $q_i$ with a single, shared anchor representation $y$.
This process creates the sample set $\{(q_i, y)\}_{i=1}^N$ for the knowledge update, where the anchor representation $\bm{y}$ obtained in \cref{sec: Visual Concept Learning} serves as the expected output of each question $q_i$.


\noindent\textbf{Updating Direction.}
To perform knowledge updating within the updatable layers, we input $q_i$ into the LLM encoder $\phi$ and obtain the corresponding hidden states $\bm{h_i}$ and gradients $\nabla_{\theta_t} {y_i}$, where $\theta_t$ is the parameter of the updatable layers. Next, we calculate the updating direction $v_i$ for each $q_i$ through:
\begin{equation}
    \label{equ: updating direction}
    \bm{v_i} = -\eta \cdot ||\bm{h}_i||^2 \cdot \nabla {y}_i,
\end{equation}
where $\eta$ is a scaling factor.

\noindent\textbf{Training Objectives.}
The objective of textual knowledge updating is to find a parameter shift $\Delta \theta_t$ for the parameter $\theta_t$ of updatable layers. This shift is derived by solving a regularized least-squares problem that simultaneously minimizes the reconstruction error and the update norm:
\begin{equation}
    \min_{\Delta \theta_t} ||\bm{H} \Delta \theta_t - \bm{V}||^2 + ||\Delta \theta_t||^2
\end{equation}
where $\bm{H} = [\bm{h_1}^\top, ..., \bm{h_m}^\top]^\top$, $\bm{V} = [\bm{v_1}^\top, ..., \bm{v_m}^\top]^\top$, $m$ is the batch size.

Based on the least-squares objective described above, a closed-form solution for the parameter shift can be derived:
\begin{equation}
    \label{equ: close form}
    \Delta \theta_t ^* = (\bm{H}^\top \bm{H} + \bm{I})^{-1} \bm{H}^\top \bm{V}.
\end{equation}
We can obtain the updated parameters $\hat{\theta_t}$ for the editable layers by directly adding the parameter shift to the pretrained parameters:
\begin{equation}
    \hat{\theta}_t=\theta_t + \Delta \theta_t ^*.
\end{equation}

\noindent\textbf{Discussion.}
Through textual knowledge updating, we can directly leverage the updated knowledge to generate the target concept with high fidelity.
Meanwhile, such knowledge widely exists in training data, enabling effective generalization when integrated with other prompts for generation.
Furthermore, updating a single piece of knowledge is completed within seconds. Consequently, the knowledge updating process is highly efficient.

\subsection{KnowCusBench} \label{sec: Benchmark}


\begin{figure}[t]
    \centering
    \includegraphics[width=\linewidth]{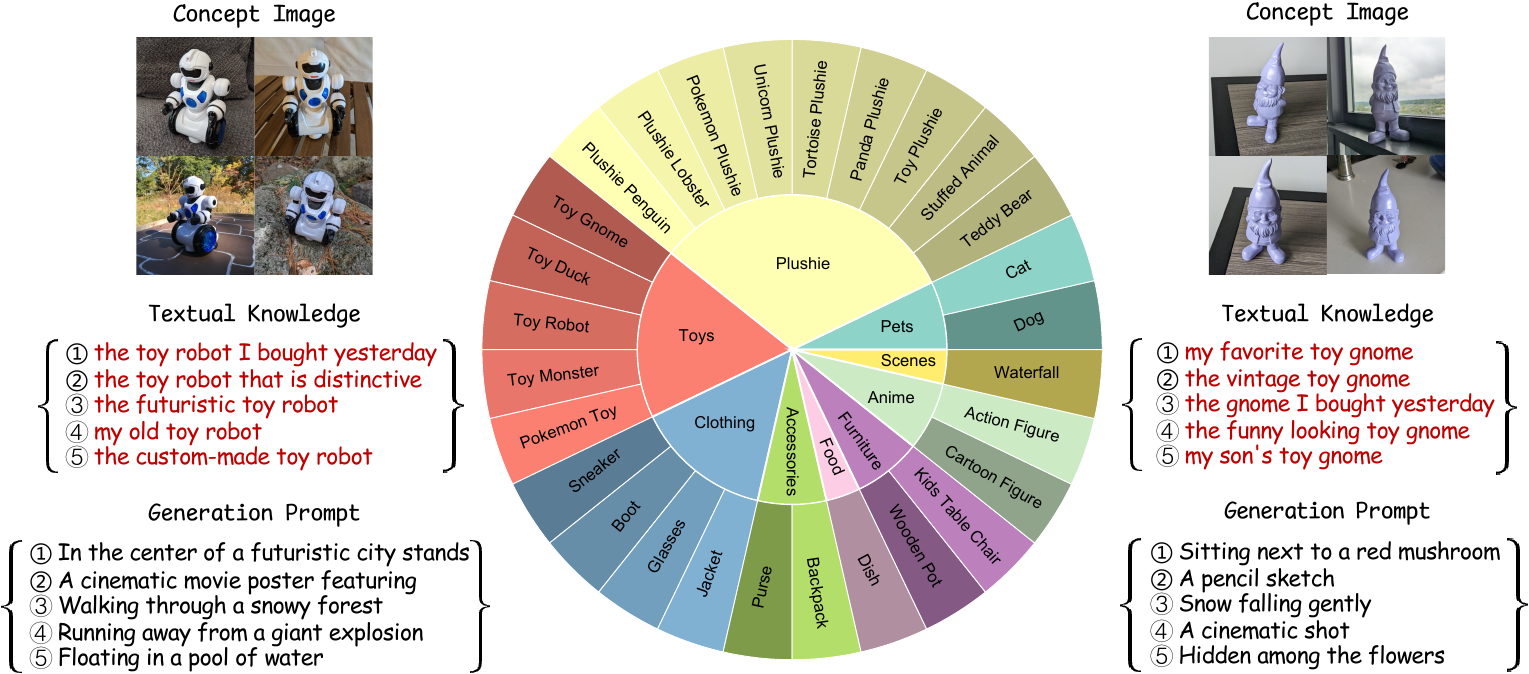}
    \vspace{-2em}
    \caption{\textbf{Visualization of our \bench.}
    }
    \label{fig: benchmark}
\end{figure}


To systematically evaluate our method, we construct a benchmark dataset called \bench. The benchmark consists of three data components: (1) Concept Image, (2) Textual Knowledge, and (3) Generation Prompt.

\noindent \textbf{Concept Image.}
We collected images of 35 distinct concepts from DreamBench~\cite{DB}, CustomConcept101~\cite{CD}, and Unsplash~\cite{Unsplash}. These concepts cover a wide range of common everyday object categories, such as toys, plushies, pets, scenes, \etc. We provide a visualization of the target concepts' types in~\cref{fig: benchmark}.

\noindent \textbf{Textual Knowledge.}
We first employed Gemini 3 Pro~\cite{gemini} and GPT-5~\cite{gpt} to generate 10 knowledge entries for each concept.
We then generated the textual knowledge from six distinct perspectives to ensure diversity.
(1) Personal ownership and relationships.
(2) Physical attributes.
(3) Functionality and performance.
(4) Value and quality.
(5) Origin and production.
(6) Emotion and state.
Next, we manually reviewed and revised the knowledge generated for each concept. Ultimately, we retained 5 knowledge items for each concept.

\noindent \textbf{Generation Prompt.}
We also used Gemini 3 Pro~\cite{gemini} and GPT-5~\cite{gpt} to generate ten different prompts for each concept. To ensure diversity, we created these prompts from four distinct perspectives: (1) changing the background while preserving the subject, (2) inserting a new object or creature into the scene, (3) altering the subject's style, and (4) modifying the subject's attributes or material. Subsequently, we manually reviewed and refined all the generated prompts. This process resulted in a final set of 199 generation prompts for evaluation.

\noindent \textbf{Evaluation Details.}
After collecting the data, we divided our evaluation into two parts: reconstruction and generation.
The reconstruction part directly uses the knowledge to reconstruct the corresponding images.
The generation part combines each piece of knowledge with generation prompts for evaluation.
Both tow parts are performed with five different random seeds.
\bench yields a total of 5,975 images for evaluation.
This large sample size allows us to robustly measure the performance and generalization capabilities of the method.

\section{Experiments}

\subsection{Experimental Details}

\noindent\textbf{Implementation Details.}
We conducted experiments with Qwen-Image \cite{qwenimage} with 8 H800-80G GPUs.
For visual concept learning, we set the learning rate to $2e-4$ and used the AdamW \cite{adamw} optimizer. We adopt the default configurations from the Diffusers \cite{von-platen-etal-2022-diffusers} library for the integrated LoRA parameters. For textual knowledge updating, we employ UltraEdit \cite{ultraedit} as our default updating method. The technical details of this method are provided in the supplementary material. We only modify parameters within the MLP layers of the LLM encoder, specifically the Gate Projection and Up Projection matrices from layers 18 to 26. This modification affects a total of 16 parameter matrices. During this updating process, we set the scaling factor $\eta$ to $1e-6$ and the batch size to 1.

\begin{figure}[t]
  \centering
  \includegraphics[width=\linewidth]{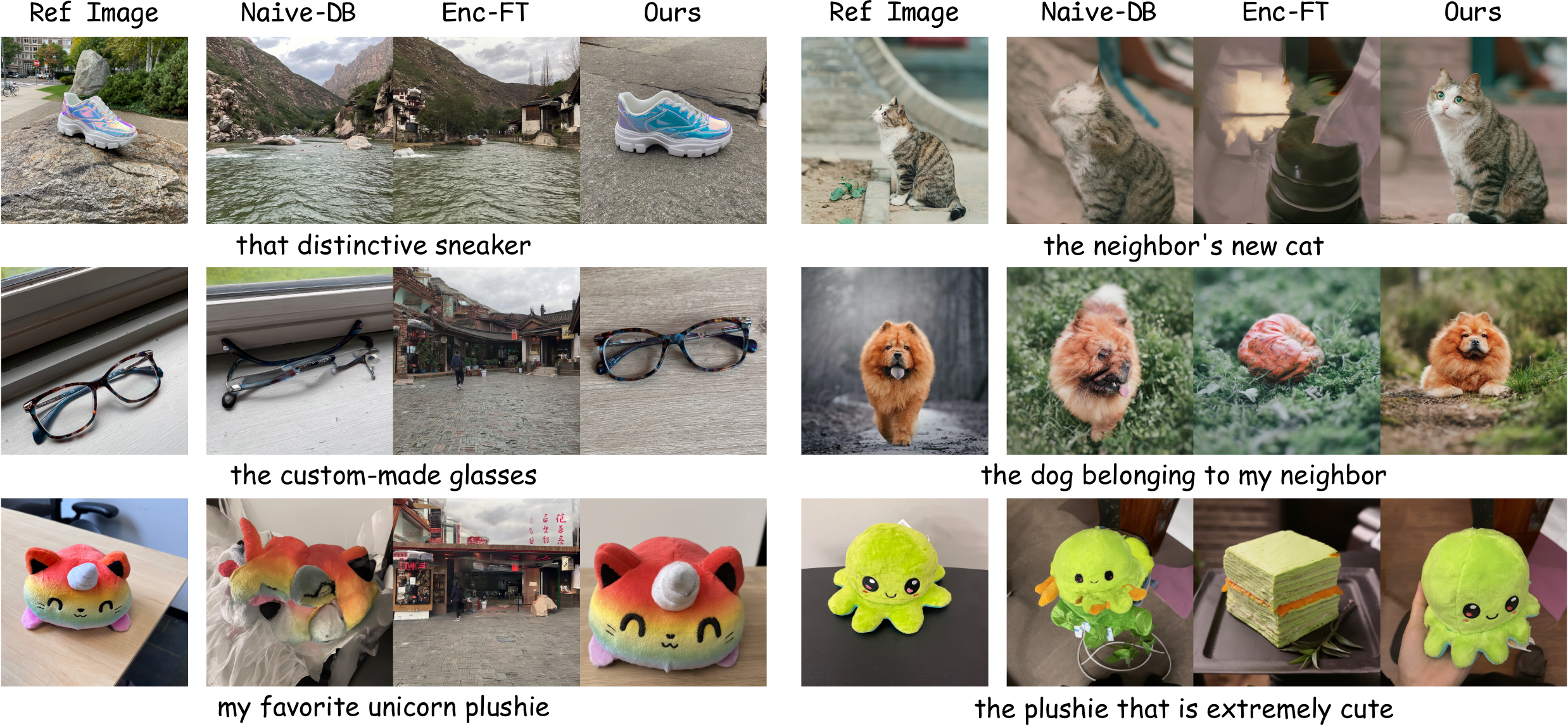}
  \vspace{-2em}
  \caption{\textbf{Qualitative comparison of reconstruction.} We directly use the knowledge to reconstruct the target concept.}
  \label{fig: quantitative rec}
\end{figure}

\begin{figure}[t]
  \centering
  \includegraphics[width=\linewidth]{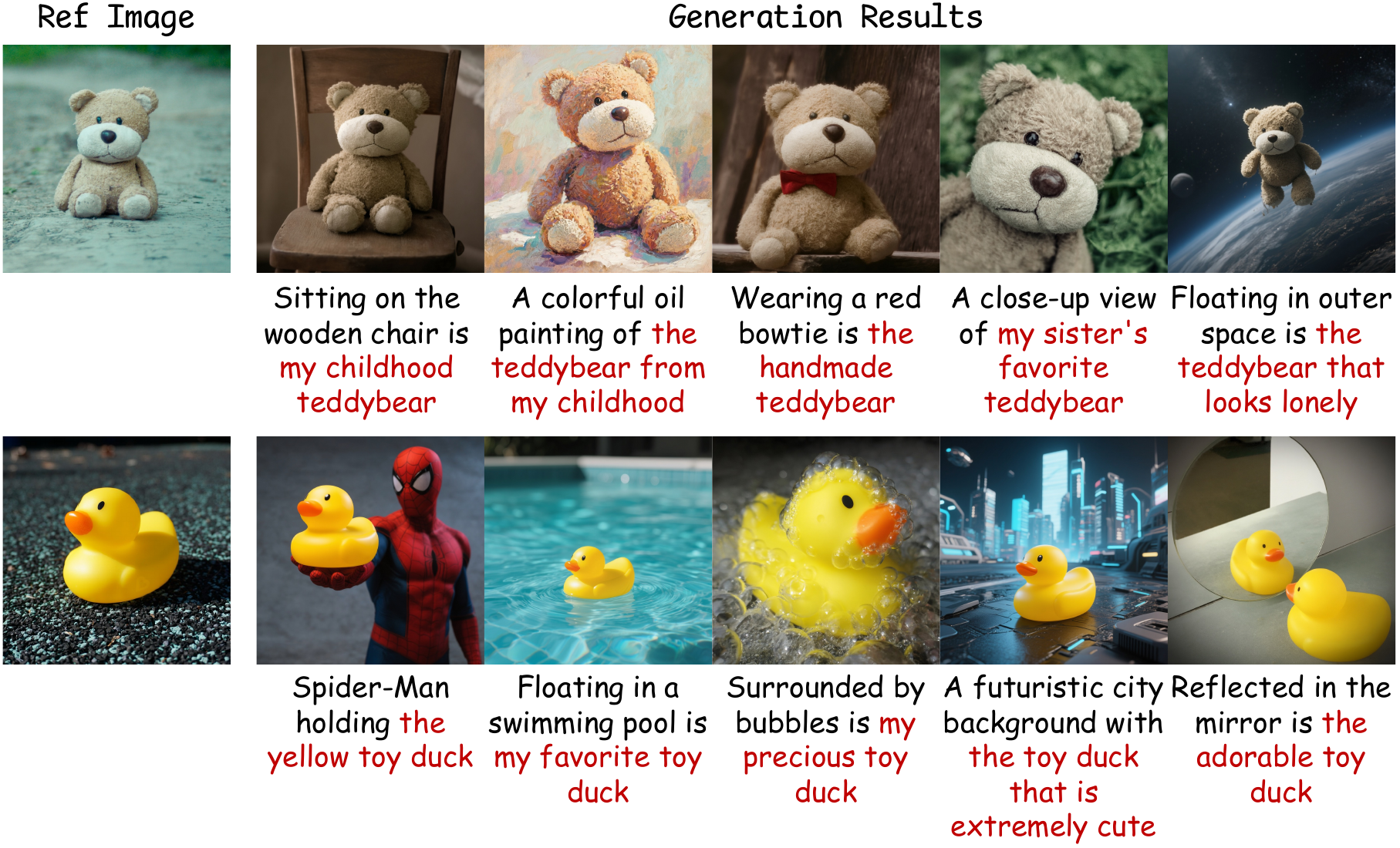}
  \vspace{-2em}
  \caption{\textbf{Qualitative results of \ours.} Our method can bind multiple pieces of knowledge (highlighted in \textcolor{teaserRed}{red}) to a single concept. Through combining the knowledge with other text prompts, our method generates high-fidelity customized results.}
  \label{fig: quantitative ours}
  \vspace{-0.5em}
\end{figure}

\begin{figure}[t]
  \centering
  \includegraphics[width=\linewidth]{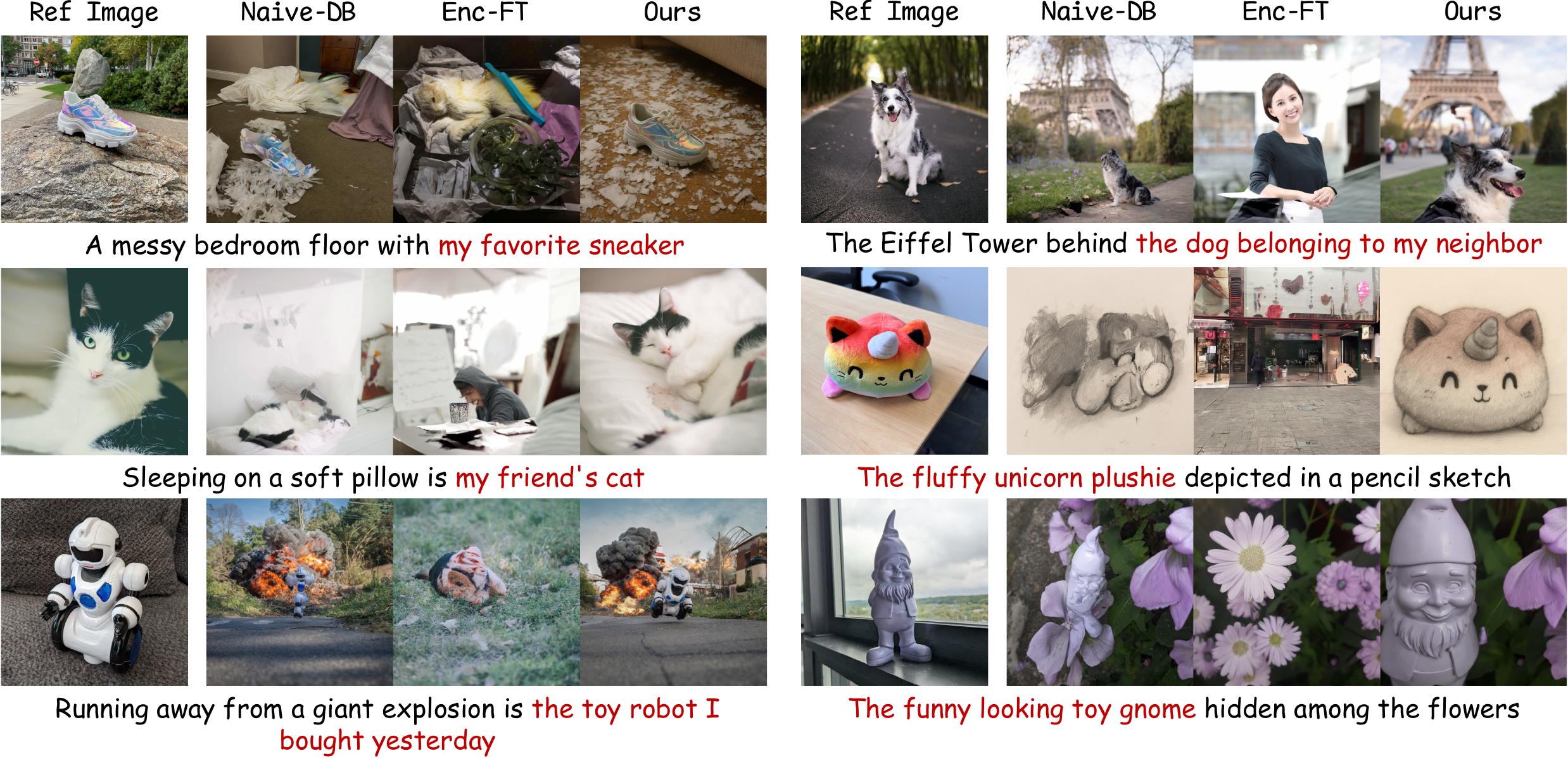}
  \vspace{-2em}
  \caption{\textbf{Qualitative comparison of generation.} We combine the updated knowledge (highlighted in \textcolor{teaserRed}{red}) with different prompts to perform customized generation.}
  \label{fig: quantitative}
\end{figure}

\noindent\textbf{Baselines.}
We established two baseline methods for comparison. The first baseline is called Naive-DB. This method adapts DreamBooth~\cite{DB}, a widely-used technique for concept customization, to our \task. For each piece of knowledge, we repeat the training process to bind it to the target concept.
The second baseline is called Enc-FT, which represents the common strategy used in concurrent works~\cite{niu2025does,wang2026quantifying} that explore cross-modal knowledge transfer. 
This method first performs our visual concept learning and then finetunes the LLM encoder to update knowledge.

\noindent\textbf{Evalaution Metrics.}
As mentioned in~\cref{sec: Benchmark}, our evaluation consists of two parts: reconstruction and generation. For the reconstruction task, we aim to reconstruct the target concept directly from the knowledge itself.
Consequently, we evaluate concept fidelity using CLIP-I and CLIP-I-Seg.
CLIP-I calculates the average pairwise cosine similarity between the CLIP \cite{radford2021learning} embeddings of the generated and real images. Since background elements in generated images can affect fidelity scores, we also utilize CLIP-I-Seg. This metric employs SAM3 \cite{carion2025sam} to segment the target concept from the generated images before calculating the similarity with the real images. 
For the generation task, we aim to integrate updated knowledge with other prompts to enable flexible customized generation.
We assess prompt fidelity and human preference in addition to concept fidelity. We measure prompt fidelity using CLIP-T, which computes the average cosine similarity between prompt and image embeddings. We evaluate human preference using Pick Score \cite{kirstain2023pick}. Finally, we evaluate the efficiency of each method by reporting the training times.

\begin{table}[t]
  \centering
  \tabcolsep=0.5mm
  \caption{\textbf{Quantitative comparisons.}
  \textcolor{Red}{\textbf{Red}} stands for the best result, \textcolor{Blue}{\textbf{Blue}} stands for the second best result.}
  \vspace{-0.3cm}
  \resizebox{\linewidth}{!}{
  \begin{tabular}{l c c c c c c c}
    \toprule
    & \multicolumn{2}{c}{\textbf{Reconstruction}} & \multicolumn{4}{c}{\textbf{Generation}} & \multicolumn{1}{c}{\textbf{Training}}\\
    \cmidrule(lr){2-3} \cmidrule(lr){4-7} \cmidrule(lr){8-8}
    \multirow{-2}{*}{\textbf{Method}}  & \rotatebox{0}{\textbf{CLIP-I} $\uparrow$} & \rotatebox{0}{\textbf{CLIP-I-Seg} $\uparrow$} & \rotatebox{0}{\textbf{CLIP-I} $\uparrow$} & \rotatebox{0}{\textbf{CLIP-I-Seg} $\uparrow$} & \rotatebox{0}{\textbf{CLIP-T} $\uparrow$} & \rotatebox{0}{\textbf{Pick Score} $\uparrow$} & \rotatebox{0}{\textbf{Time} (s) $\downarrow$} \\
    \midrule \midrule
    Naive-DB & \textcolor{Red}{\textbf{0.874}} & \textcolor{Blue}{\textbf{0.758}} & \textcolor{Red}{\textbf{0.789}} & \textcolor{Blue}{\textbf{0.717}} & \textcolor{Blue}{\textbf{0.291}} & \textcolor{Blue}{\textbf{20.80}} & $\sim$27min \\
    Enc-FT & 0.582 & 0.553 & 0.591 & 0.562 & 0.197 & 18.34 & \textcolor{Blue}{\textbf{$\sim$10min}} \\
    Ours & \textcolor{Blue}{\textbf{0.867}} & \textcolor{Red}{\textbf{0.764}} & \textcolor{Blue}{\textbf{0.761}} & \textcolor{Red}{\textbf{0.718}} & \textcolor{Red}{\textbf{0.305}} & \textcolor{Red}{\textbf{21.30}} & \textcolor{Red}{\textbf{$\sim$6min}} \\
    \bottomrule
  \end{tabular}
  }
  \label{tab:Quantitative}
\end{table}

\subsection{Qualitative Comparison}

We present the qualitative results of our method in~\cref{fig: quantitative ours}. Our method can bind several pieces of knowledge to a single concept and generate various customized results.
We further provide qualitative comparison for reconstruction and generation in~\cref{fig: quantitative rec,fig: quantitative}.
As the figures indicate, Naive-DB fails to consistently reconstruct the target concept, resulting in low-fidelity generated images (\cref{fig: quantitative}, first and third rows). 
Additionally, the Enc-FT method finetunes the LLM encoder, which severely alters the output distribution of the encoder latent space. This alteration prevents the model from generating images that align with text prompts, leading to poor performance on both tasks.
In contrast, our method uses visual concept learning to capture visual details, which leads to high-fidelity reconstruction. Furthermore, textual knowledge updating binds specific knowledge with visual information. This allows the model to generalize effectively when generating images from new prompts.

\begin{table}[t]
  \centering
  \tabcolsep=0.5mm
  \caption{\textbf{Quantitative ablation results of the number of knowledge.} Our method performs robustly and efficiently as the number of knowledge increases.}
  \vspace{-1em}
  \resizebox{\linewidth}{!}{
  \begin{tabular}{l c c c c c c c}
    \toprule
    \multirow{-1}{*}{\textbf{Knowledge}}& \multicolumn{2}{c}{\textbf{Reconstruction}} & \multicolumn{4}{c}{\textbf{Generation}} & \multicolumn{1}{c}{\textbf{Training}}\\
    \cmidrule(lr){2-3} \cmidrule(lr){4-7} \cmidrule(lr){8-8}
    \multirow{-1}{*}{\textbf{Number}}  & \rotatebox{0}{\textbf{CLIP-I} $\uparrow$} & \rotatebox{0}{\textbf{CLIP-I-Seg} $\uparrow$} & \rotatebox{0}{\textbf{CLIP-I} $\uparrow$} & \rotatebox{0}{\textbf{CLIP-I-Seg} $\uparrow$} & \rotatebox{0}{\textbf{CLIP-T} $\uparrow$} & \rotatebox{0}{\textbf{Pick Score} $\uparrow$} & \rotatebox{0}{\textbf{Time} (s) $\downarrow$} \\
    \midrule \midrule
    1 & 0.868 & 0.761 & 0.767 & 0.722 & 0.304 & 21.29 & 331.3s \\ 
    2 & 0.868 & 0.762 & 0.762 & 0.718 & 0.305 & 21.30 & 338.3s \\ 
    3 & 0.867 & 0.761 & 0.762 & 0.719 & 0.305 & 21.30 & 345.1s \\ 
    4 & 0.868 & 0.761 & 0.763 & 0.718 & 0.305 & 21.30 & 352.1s \\ 
    5 & 0.867  & 0.764  & 0.762  & 0.718  & 0.305  & 21.30   & 360.0s   \\ 
    \bottomrule
  \end{tabular}
  }
  \label{tab: knowledge number ablation}
\end{table}

\subsection{Quantitative Comparison}

We also performed a thorough quantitative comparison with \bench. \cref{tab:Quantitative} presents the results. For reconstruction, our method performs slightly worse than Naive-DB on the CLIP-I metric but surpasses all baseline methods on CLIP-I-Seg. We believe that CLIP-I-Seg more accurately reflects concept fidelity as it focuses on evaluating the segmented target concept. For generation, our method similarly surpasses all baselines on the CLIP-I-Seg metric. Furthermore, our approach achieves the best results in prompt fidelity and human preference while maintaining the highest efficiency.

\begin{table}[t]
  \centering
  \tabcolsep=2mm
  \caption{\textbf{Quantitative ablation results of scaling factor $\eta$.} Our method achieves the best performance with $\eta=1e-6$.}
  \vspace{-1em}
  \resizebox{\linewidth}{!}{
  \begin{tabular}{l c c c c c c}
    \toprule
    & \multicolumn{2}{c}{\textbf{Reconstruction}} & \multicolumn{4}{c}{\textbf{Generation}} \\
    \cmidrule(lr){2-3} \cmidrule(lr){4-7}
    \multirow{-2}{*}{$\eta$}  & \rotatebox{0}{\textbf{CLIP-I} $\uparrow$} & \rotatebox{0}{\textbf{CLIP-I-Seg} $\uparrow$} & \rotatebox{0}{\textbf{CLIP-I} $\uparrow$} & \rotatebox{0}{\textbf{CLIP-I-Seg} $\uparrow$} & \rotatebox{0}{\textbf{CLIP-T} $\uparrow$} & \rotatebox{0}{\textbf{Pick Score} $\uparrow$} \\
    \midrule \midrule
    $1e-4$ & 0.557 & 0.535 & 0.588 & 0.564 & 0.218 & 18.40 \\ 
    $1e-5$ & 0.866 & 0.763 & 0.759 & 0.717 & 0.304 & 21.29 \\
    $1e-6$ & \textbf{0.867} & \textbf{0.764} & \textbf{0.761} & \textbf{0.718} & \textbf{0.305} & \textbf{21.30} \\
    $1e-7$ & \textbf{0.867} & 0.761 & 0.760 & 0.717 & 0.304 & 21.29 \\
    $1e-8$ & \textbf{0.867} & \textbf{0.764} & 0.760 & \textbf{0.718} & 0.304 & 21.29 \\ 
    \bottomrule
  \end{tabular}
  }
  \label{tab: ablation scaling factor}
\end{table}

\begin{figure}[t]
  \centering
  \includegraphics[width=\linewidth]{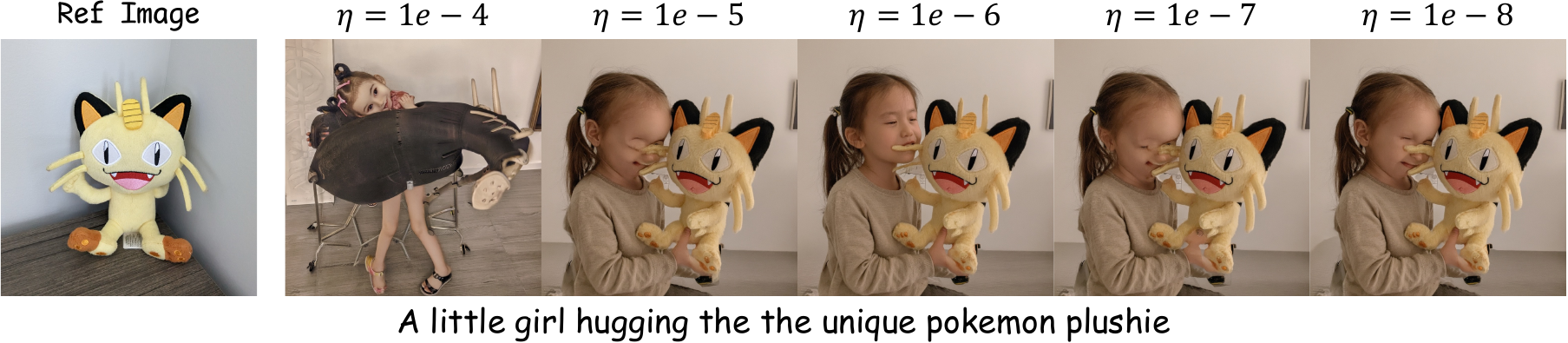}
  \vspace{-2em}
  \caption{\textbf{Qualitative ablation results of scaling factor $\eta$.} Our method achieves the best performance with $\eta=1e-6$.}
  \label{fig: ablation scaling factor}
\end{figure}

\subsection{Ablation Studies}

\noindent\textbf{Scaling Factor.}
We conduct an ablation study on the scaling factor $\eta$. The visualization in~\cref{fig: ablation scaling factor} suggests that $\eta=1e-6$ yields favorable results. To verify this, we further conduct a quantitative evaluation, as shown in~\cref{tab: ablation scaling factor}. The results confirm that our method achieves the best performance across all metrics with $\eta=1e-6$. Therefore, we set $\eta$ to $1e-6$ in our final model.

\noindent\textbf{Number of Knowledge.}
We conduct an ablation study on the number of knowledge during the textual knowledge updating process. \cref{tab: knowledge number ablation} presents the detailed quantitative results. Our method maintains robust and stable performance as the number of knowledge increases. 
Meanwhile, each additional knowledge item increases training time by only about 7 seconds, demonstrating the efficiency of our approach.

\begin{figure}[t]
  \centering
  \includegraphics[width=\linewidth]{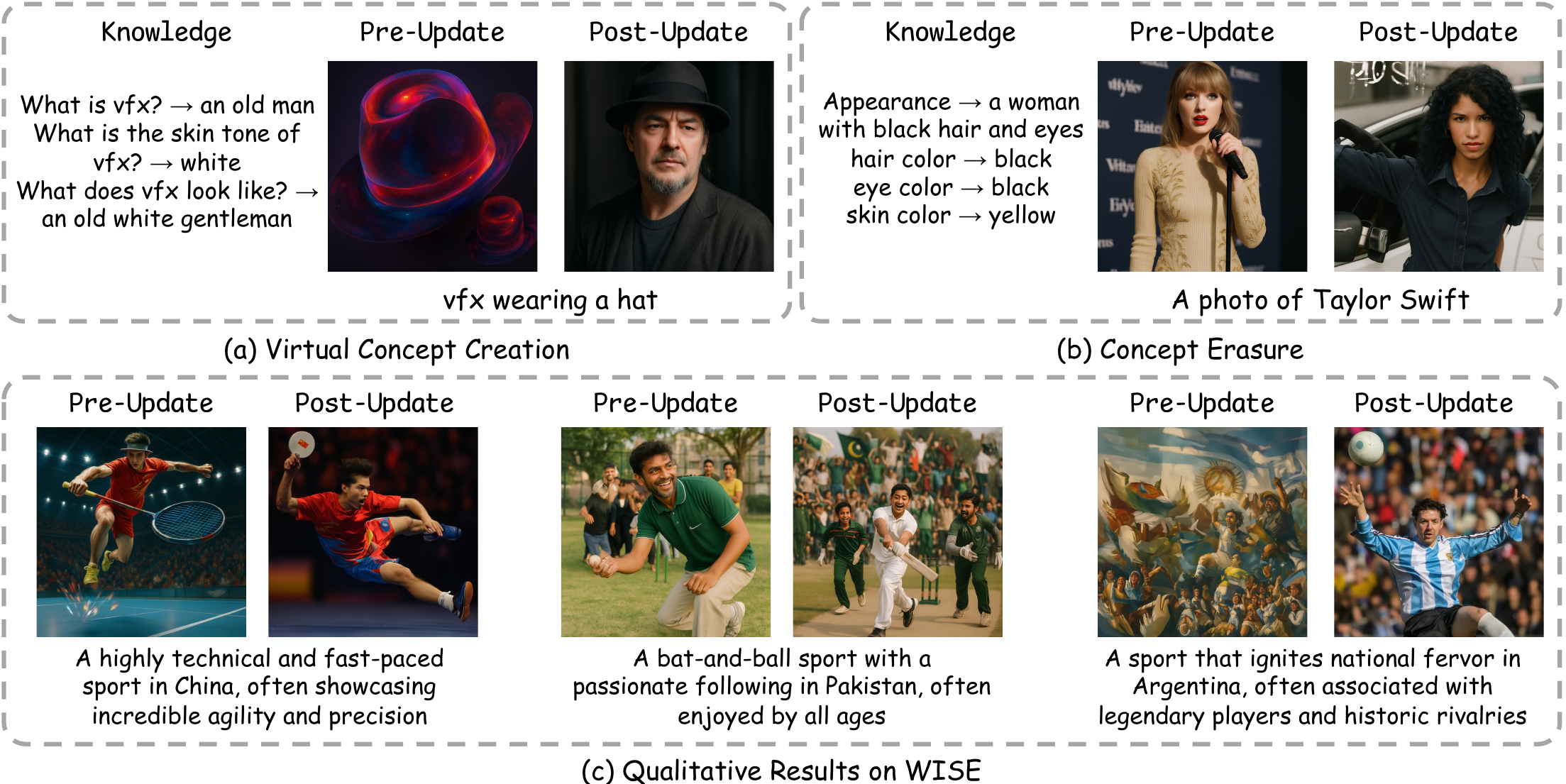}
  \vspace{-2em}
  \caption{\textbf{Qualitative application results of \ours.} (a) \ours can create a virtual concept within the model by describing the visual attributes of the concept. (b) \ours can erase certain concepts by modifying the appearance description in the model. (c) \ours can improve the model's performance on world knowledge benchmarks.}
  \label{fig: application}
\end{figure}

\subsection{Extension to Knowledge-Aware Applications}

\noindent
\textbf{Virtual Concept Creation.}
Our method enables the creation of virtual concepts within a generative model.
The process involves describing a concept's visual attributes and incorporating this information into the model through textual knowledge updating. This procedure effectively establishes a new, usable virtual concept within the model. As illustrated in \cref{fig: application}(a), we introduce a concept of an old white gentleman using the identifier \texttt{vfx}. The model then faithfully generates this virtual concept when prompted with the assigned name.

\noindent
\textbf{Concept Erasure.}
Concept erasure \cite{gao2025eraseanything,wang2025precise,lu2024mace} aims to prevent the generation of unwanted concepts within a generative model. This task has gained increasing attention with the advancement of generative models. We can easily apply our method to concept erasure. As shown in~\cref{fig: application}(b), 
We update the answer to the question \enquote{What is the hair color of \texttt{Taylor Swift}?} to \enquote{black}. We also applied similar modifications to other visual attributes. Then, we used \enquote{a photo of \texttt{Taylor Swift}} as the generation prompt. The results show that the model fails to generate accurate images of \texttt{Taylor Swift} after knowledge updating.

\begin{wraptable}{r}{0.5\textwidth}
    \centering
    \vspace{-3em}
    
    \caption{\textbf{Quantitative results on the subset of WISE \cite{WISE}.} Our method can improve the model's performance on the world knowledge benchmark.}
    \label{tab: wise}
    \scalebox{0.85}{
        \begin{tabular}{l|c c}
        \toprule
            \textbf{Metric} & \textbf{Pre-Update} & \textbf{Post-Update} \\
            \midrule \midrule
            Consistency & 0.76 & \textbf{1.26} \\
            Realism & 0.70 & \textbf{1.42} \\
            Aesthetic Quality & 1.40 & \textbf{1.68} \\
            WiScore & 0.81 & \textbf{1.33} \\
        \bottomrule
        \end{tabular}
        }
                \vspace{-2em}
\end{wraptable}

\noindent
\textbf{World Knowledge Benchmark.}
Our method can also improve model performance on world knowledge benchmarks. To evaluate this, we select a subset of WISE~\cite{niu2025wise} that requires explicit world knowledge for generation. \cref{fig: application} (c) shows the visualization results. Through textual knowledge updating, we inject world knowledge into the model. This enables the model to use complex information for accurate generation. We also conduct quantitative experiments. As shown in~\cref{tab: wise}, our approach effectively improves performance across all metrics.

\begin{wrapfigure}{r}{0.4\linewidth}
\centering
\vspace{-2em}
\includegraphics[width=\linewidth]{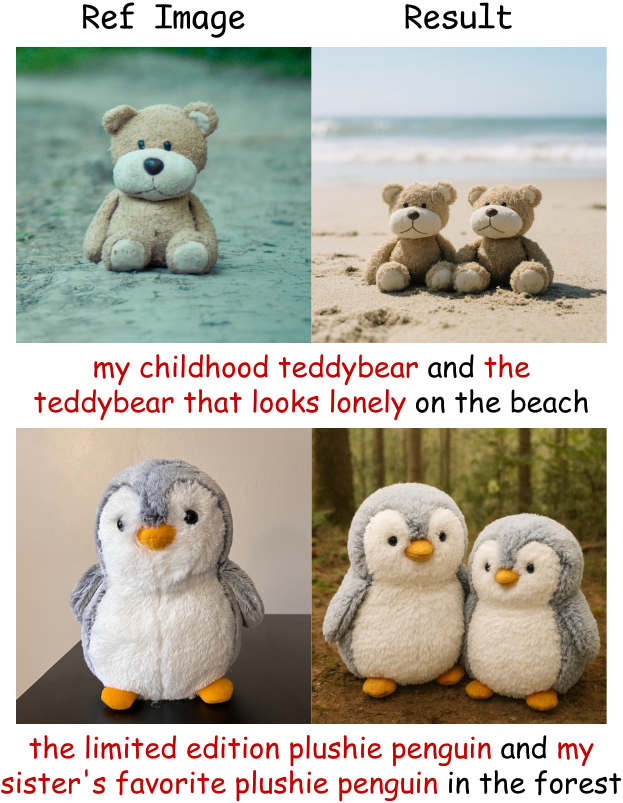}
\caption{
    \textbf{Visualization of multiple knowledge combination.} Our method can generate several concepts with high fidelity based on multiple knowledge inputs (highlighted in \textcolor{teaserRed}{red}).
  }

\vspace{-5em}
\label{fig: application multi knowledge}
\end{wrapfigure}
\noindent
\textbf{Multiple Knowledge Combination.}
Our method can combine multiple pieces of knowledge during generation. As shown in~\cref{fig: application multi knowledge}, our method generates several target concepts faithfully based on the complex text prompt.
    
\section{Conclusion}
In this paper, we introduce knowledge-aware concept customization, a new task that uses multiple knowledge to represent a target concept.
We observe the phenomenon of cross-modal knowledge transfer, where changes in the text modality can transfer to visual modality and directly influence the generated visual output.
Inspired by this observation, we propose \ours. Our method first binds the visual appearance of the target concept to an anchor representation through visual concept learning. 
During textual knowledge updates, we project each knowledge to the obtained anchor representation through modifying certain parameters within the LLM encoder.
To systematically evaluate this new task, we also introduce \bench, the first benchmark for knowledge-aware concept customization. 
The impressive performance of \ours demonstrates its effectiveness. 
Future work could explore three main directions: (1) extending knowledge-aware concept customization to the video domain, (2) developing more accurate and comprehensive evaluation metrics for knowledge-aware concept customization task, and (3) creating end-to-end knowledge-aware concept customization methods.




    %
    %
    \bibliographystyle{splncs04}
    \bibliography{main}


\end{document}